%% file: 0.main.tex
\documentclass[runningheads]{llncs}
\usepackage{amsmath}
\usepackage[T1]{fontenc}
\usepackage{graphicx}
\usepackage{amsfonts}
\usepackage{color}
\begin{document}
\makeatletter
\newcommand\thanksstar{\raisebox{0.7ex}{\scriptsize\star}}
\makeatother
\title{Structured Dialogue System for Mental Health: An LLM Chatbot Leveraging the PM$^+$ Guidelines}
\titlerunning{Structured Dialogue System for Mental Health}
%

\author{Yixiang Chen\inst{1,2,3}
\and
Xinyu Zhang\inst{1,4}
\and
Jinran Wang\inst{1,5}
\and
Xurong Xie\inst{3}
\and
Nan Yan\inst{1}
\and
Hui Chen\inst{3}
\and
Lan Wang\inst{1,\thanks{Corresponding Author}}
}
\authorrunning{Y. Chen et al.}
%
\institute{Guangdong-Hong Kong-Macao Joint Laboratory of Human-Machine Intelligence-Synergy Systems, Shenzhen Institute of Advanced Technology, Chinese Academy of Sciences, China. \\
\email{\{yx.chen7, xy.zhang14, jr.wang2, nan.yan, lan.wang\}@siat.ac.cn}
\and
University of Chinese Academy of Sciences, Beijing, China
\and
Institute of Software, Chinese Academy of Sciences Beijing, China \\
\email{\{xurong,chenhui\}@iscas.ac.cn}
\and
East China Normal University, Shanghai, China
\and
Wuhan Research Institute of Posts and Telecommunications, Wuhan, China
}
\maketitle              
\begin{abstract}
The Structured Dialogue System, referred to as SuDoSys, is an innovative Large Language Model (LLM)-based chatbot designed to provide psychological counseling. SuDoSys leverages the World Health Organization (WHO)'s Problem Management Plus (PM+) guidelines to deliver stage-aware multi-turn dialogues. Existing methods for employing an LLM in multi-turn psychological counseling typically involve direct fine-tuning using generated dialogues, often neglecting the dynamic stage shifts of counseling sessions. Unlike previous approaches, SuDoSys considers the different stages of counseling and stores essential information throughout the counseling process, ensuring coherent and directed conversations. The system employs an LLM, a stage-aware instruction generator, a response unpacker, a topic database, and a stage controller to maintain dialogue flow. In addition, we propose a novel technique that simulates counseling clients to interact with the evaluated system and evaluate its performance automatically. When assessed using both objective and subjective evaluations, SuDoSys demonstrates its effectiveness in generating logically coherent responses. The system's code and program scripts for evaluation are open-sourced\footnote{\url{https://github.com/EthanLifeGreat/SuDoSys}}.

\end{abstract}
\input{1.intro}
\input{3.method}
\input{4.experiments}
\input{5.conclusion}

\begin{credits}
\section{\ackname}
This work was supported by National Natural Science Foundation of China (NSFC U23B2018, 62106255) and  Shenzhen Science and Technology Program (No. KQTD2020\\0820113106007), ShenZhen Fundamental Research Program (JCYJ20220818101411025), and Youth Innovation Promotion Association CAS Grant 2023119.



We would like to express our sincere gratitude to Rennan Wang for her invaluable assistance in revising and improving this manuscript. Her insightful feedback and suggestions significantly enhanced the quality of our work.

\end{credits}

\input{bib}
\end{document}

%% file: 1.intro.tex
\section{Introduction}
In today's fast-paced world, mental health has risen to the forefront of global health priorities, recognized as a critical component of overall well-being\cite{nhwmh}. However, access to professional mental health services like psychological counseling remains a challenge, particularly in regions lacking specialists. Advancements of Natural Language Processing (NLP) and Large Language Models (LLMs)\cite{zhao2023surveylargelanguagemodels} offer a promising avenue to bridge this gap. LLMs like ChatGPT\footnote{\url{https://chatgpt.com/}}, LLaMA\cite{llama,llama2}, ChatGLM\cite{chatglm} and Qwen\cite{qwen}, characterized by their vast capacity to understand instructions and generate human-like responses, have the potential to serve as empathetic listeners and psychological consultants, providing immediate assistance to help-seeking individuals.

However, LLMs must be adapted for psychological counseling, as they are primarily pre-trained for general purposes. Previous studies have utilized (real or synthetic) counseling dialogues to fine-tune pre-trained LLMs for multi-turn dialogue systems. For example, ExTES-LLaMA\cite{extes}, MEChat\cite{smile}, SoulChat\cite{soulchat}, and CPsyCounX\cite{CPsyCoun} are LLMs fine-tuned on generated multi-turn dialogues. PsyChat\cite{Psychat} is a dialogue system fine-tuned on dialogues with counselor's strategy and user's behavior labeled by human. However, these fine-tuning approaches often do not consider the different stages of counseling, resulting in dialogues that lack direction and coherence.

To address these limitations, we propose \textbf{S}tr\textbf{u}ctured \textbf{D}ial\textbf{o}gue \textbf{Sys}tem (SuDoSys), an LLM-based multi-turn dialogue system designed for stage-aware counseling. To enhance the system's intelligence in counseling, we leverage the World Health Organization(WHO)'s guidelines for structured psychological interventions—Problem Management Plus (PM+) \cite{pmp}. By adhering to the PM+’s seven-step framework for problem management (see Chapter 7 of PM+), we have meticulously crafted a series of prompts that guide SuDoSys through the entire problem management process while providing emotional support.

In this paper, we introduce SuDoSys and propose a novel automatic evaluation method utilizing counseling dialogues collected from actual PM+ interventions. We then conduct both objective and subjective evaluations of SuDoSys. The experimental results demonstrate that our approach excels in generating coherent dialogues compared to existing fine-tuning methods. Notably, all the experiments conducted in this work are in Chinese.




%% file: 3.method.tex
\section{SuDoSys and Automatic Evaluation}\label{stage_unaware}

\subsection{Stage-unaware Prompting: A Baseline Method}
An intuitive approach to priming pre-trained Large Language Models (LLMs) for the role of psychological counselors involves explicitly assigning the counselor role and responsibilities to the model prior to the commencement of the counseling session. 

However, such a stage-unaware approach overlooks the distinct stages inherent to the counseling process. Besides, LLMs are prone to exhibit suboptimal performance (such as forgetting information) as the context length increases, which can lead to a lack of coherence in the dialogue. Consequently, the stage-unaware method is employed as a baseline in this study.

\subsection{SuDoSys: A stage-aware multi-turn dialogue system}

In each stage of a psychological counseling, a counselor engages the client by inviting them to express their thoughts and responds with a strategic approach. Simultaneously, the counselor assesses whether to advance to the next stage, based on the sufficiency of the information gathered from the client and the client's readiness for the subsequent stage of discussion. Inspired by these practices, we propose a dialogue system that not only responds to the user's input appropriately, but also accounts for the progression of the conversation through different stages.

\begin{figure}[htbp]
\includegraphics[width=\textwidth]{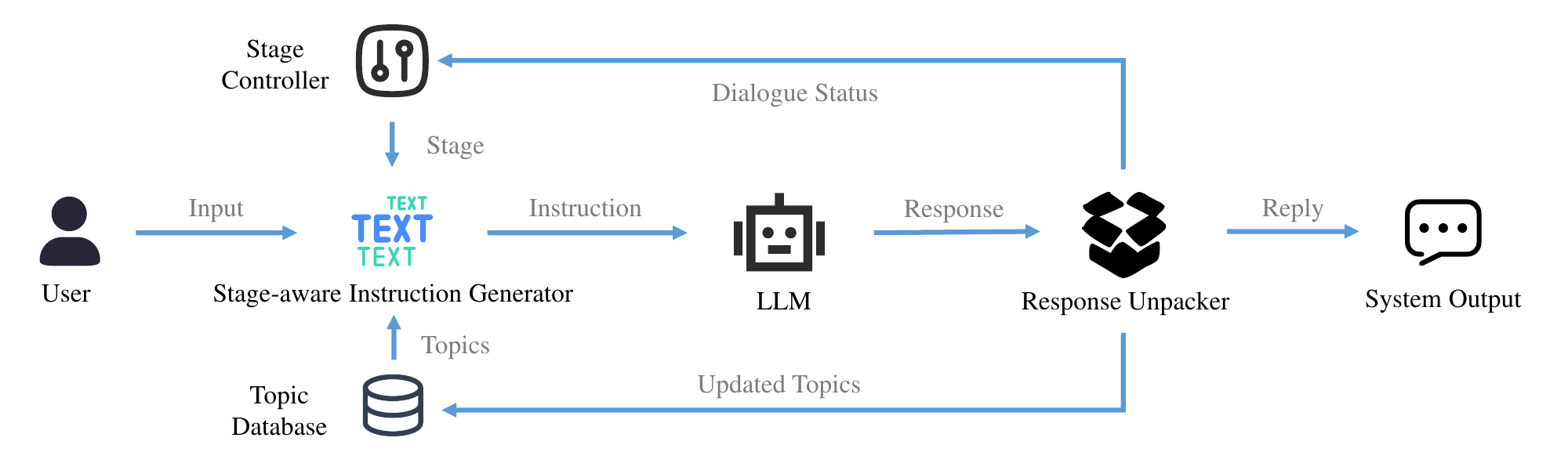}
\caption{The overview structure of SuDoSys. It consists of 5 parts: a stage controller, a stage-aware instruction generator, a topic database, a pre-trained LLM, and a response unpacker.} \label{overview}
\end{figure}

 As illustrated in Fig.~\ref{overview} SuDoSys comprises five modules: a stage controller, a stage-aware instruction generator, a topic database, a pre-trained LLM, and a response unpacker. In each turn (denoted as $t$) of the conversation, the instruction receives the current stage number $S$, the turn's user input utterance $u_t$, the conversation topics stored in all previous stages $\{\mathbb{T}_s\}_{s=1}^S$ and forms instruction $I$, where $\mathbb{T}_s=\{T_s^{k}\}_{k=1}^{K_s}$, and each $T_s^k$ is a topic that can be discussed around, and $K_s$ is the number of topics in stage $s$, as defined in topic database $D$. As instructed by $I$, the LLM then  1) extract updated conversation topics $\mathbb{T}_S^{\prime}$, 2) determine the dialogue status $c\in\{-1,0,1\}$, where $c=-1$ indicates going to the previous stage, $c=0$ indicates staying at current stage and $c=1$ indicates moving to the next stage, and 3) yield a reply $r_t$ towards the user. As soon as the topics $\mathbb{T}_S^{\prime}$ and status $c$ are produced, they are sent to the topic database and the stage controller respectively for instruction generation of the next turn. 


\subsubsection{Stage-aware Instruction Generator}
The stage-aware instruction generator is the core of SuDoSys as it sends information-retriving commands to the LLM. In stage $S$, an instruction $I$ is made up of 4 parts: 
\begin{itemize}
    \item \textbf{Stage-dependent Base Instruction} $B_S$. This is a hand-crafted directive that specifies the role and responsibilities of the LLM within the current stage of interaction. The instruction is derived from the PM+ guidelines and serves to remind the LLM of its focus during the conversation, the timing for stage shifts, and the manner in which it should respond with both empathy and professionalism. It is selected from base instruction library $\mathbb{B}$ by the generator according to $S$: $B_S=\mathbb{B}[S]$.
    \item \textbf{Stage-dependent Topics} $\{\mathbb{T}_s\}_{s=1}^S$. For each stage $s$, the stage dependent topics $\mathbb{T}_s$ represents a series of key-value pairs that define the topics to be discussed. Each key corresponds to a specific issue to be addressed, while the value provides a description of the issue. These descriptions may be revised (expanded or pruned) after each turn's discussion, with revisions applicable only to the current stage. Descriptions for topics in previous stages remain unchanged and serve as references. Similar to the base instruction, the issues for each stage are predefined based on the guidelines, and all descriptions are initialized as empty at the beginning of the dialogue. It should be distinguished with the \textit{slot pairs} proposed in Dialogue State Tracking~\cite{feng-etal-2023-towards,niu-etal-2024-enhancing}---slot values are typically constrained to a predefined set of options, whereas topic values are unrestricted descriptive sentences. 
    \item \textbf{User Input of the Turn} $u_t$. The input utterance is crucial for the LLM to comprehend the user's sentiments and gather relevant information for revising the topics above.
    \item \textbf{Response Template} $R^T_S$. This is typically represented as a Python dictionary that specifies the structure of the desired output in stage $S$, which includes the updated topics $\mathbb{T}_S^{\prime}$, the dialogue status $c\in\{-1,0,1\}$ and the user-oriented reply $r_t$.
\end{itemize}

With base instruction library $\mathbb{B}$ and Response Template $R^T_S$ stored inside, the stage-aware instruction generator $G_I$ reorganizes the values and operates as follows:
$$I = G_I(S, \{\mathbb{T}_s\}_{s=1}^S, u_t)=[\mathbb{B}[S]; \{\mathbb{T}_s\}_{s=1}^S; u_t; R^T_S],$$

{\noindent}where $I$ denotes the generated instruction; $S$ represents the current stage; $\{\mathbb{T}_s\}_{s=1}^S$ is the set of topics for stages starting from one up to the current number $S$; $u_t$ is user's input utterance; and the square brackets $[\cdot ; \cdot]$ means direct text concatenation. Examples of the instructions are available in our open-source project.

\subsubsection{LLM and Response Unpacker}
The decoder-only LLM is the core of the system functioning because it is responsible for comprehending user's input, extracting new topics $\mathbb{T}_S^{\prime}$, and generating output $r_t$ and dialogue status $c$. The text-formatted output of the LLM, denoted as $o$, is expected to contain and only contain the values above. However, due to the lack of Human Preference Alignment Training\cite{RRHF,Training}, the LLMs sometimes fail to generate responses that are pure JSON and contain exactly the dictionary keys as required. For instance, the LLM might generate incompatible quotation marks or provide additional explanations of the JSON, making the output unable to be parsed. In such cases, the response unpacker is responsible for fixing the format errors and removing the redundant messages, and parse the revised output to the values of $\mathbb{T}_S^{\prime}$, $r_t$ and $c$. As for more complex situations where the unpacker cannot handle, the LLM is required to regenerate the response.
The LLM and the response unpacker together, denoted as $F_{LLM}$, function as follows:
$$\mathbb{T}_S^{\prime}, r_t, c=F_{LLM}(I),$$
{\noindent}where $\mathbb{T}_S^{\prime}$ is the updated topics for stage $S$; $r_t$ is the reply towards user; $c$ dialogue status; and $I$ is the instruction input to the LLM. It is worth noting that, theoretically, the LLM in the SuDoSys can be any pre-trained LLM, and the processing speed of function $F_{LLM}$ is positively correlated with the degree of the LLM's alignments with human preferences.

\subsubsection{Stage Controller}
The stage controller changes the stage number $s$ according to the $c$ dialogue status:
$$ s:=s+c,$$
{\noindent}where $c\in\{-1,0,1\}$ is the dialogue status.

\subsubsection{Topic Database}
The topic database stores the aforementioned topics for each chatting stages, denoted as $\{\mathbb{T}_s\}_{s=1}^N$ where $N$ is the total number of stages in the dialogue. During each stage $S$, the database updates only the topics of the stage:
$$ \mathbb{T}_S:=\mathbb{T}_S^\prime,$$
{\noindent}but it sends to the instruction generator all topics of previous stages(including the current one):  $\{\mathbb{T}_s\}_{s=1}^S$.

\subsection{Automatic Evaluation Method for Stage-aware Multi-turn Dialogue Systems}

To automatically evaluate the AI-generated responses, many researches\cite{lin-chen-2023-llm,liu-etal-2023-g} employ GPT-4 to assess the quality of such responses.
In the context of psychological counseling, existing methods\cite{soulchat,smile,Psychat,chatcounselor} for evaluating multi-turn dialogue systems often focus on comparing single-round outputs between/among different models given the same dialogue history. However, these approaches cannot be directly applied to evaluate a stage-aware dialogue system. This is because a randomly selected dialogue history may diverge significantly from a structured dialogue, thereby rendering the comparison between models less meaningful.

\begin{figure}[htbp]
\includegraphics[width=\textwidth]{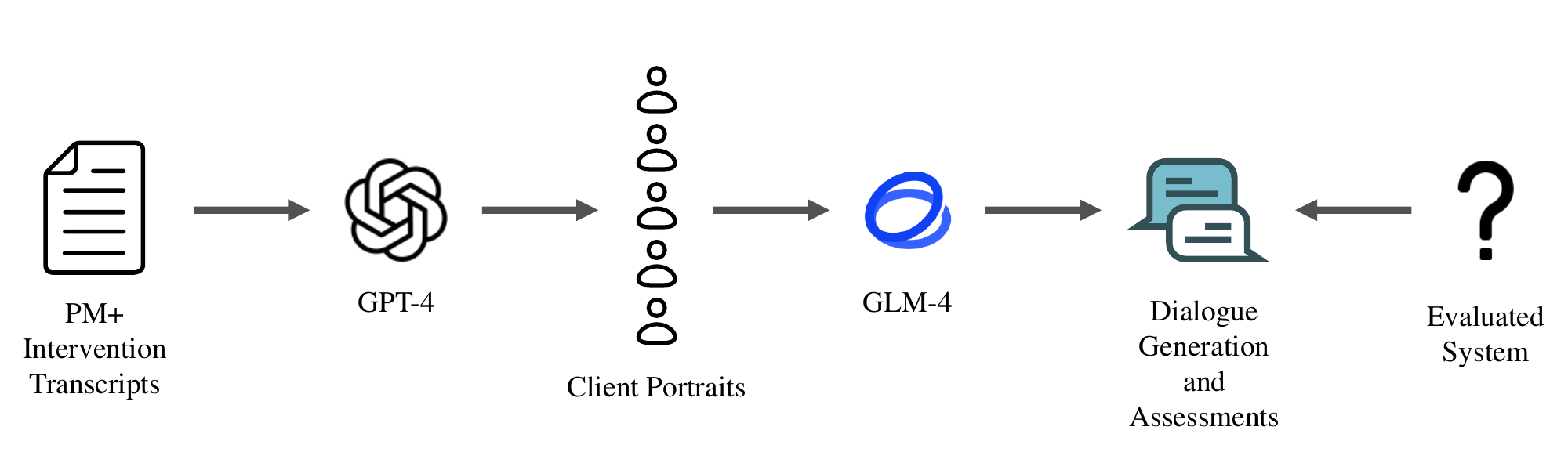}
\caption{The overview structure of our proposed evaluation method. We make use of GPT-4 to extract portraits of actual PM+ clients. The GLM-4 model then plays the role of these clients and make conversations with dialogue systems to generate dialogues for evaluation.} \label{evaluation}
\end{figure}

To address such issues, we propose an automatic evaluation method for assessing the quality of multi-turn responses. As illustrated in Fig. \ref{evaluation}, we first employ GPT-4 to extract client portraits from dialogues transcribed from raw PM+ intervention recordings. Subsequently, we prompt the GLM-4\cite{chatglm} model to act as these clients and engage in conversations with the evaluated systems to generate dialogues that will undergo further assessments. Details of the components in the figure are as follows:
\begin{itemize}
\item \textbf{PM+ Intervention Transcripts}. These are 148 dialogues transcribed from recordings of psychological counseling sessions in Chinese, in which counselors apply the PM+ intervention guidelines. Each session has a duration of approximately 90 minutes. Each dialogue comprises approximately 20,000 Chinese characters and encompasses over 100 turns of utterances.
\item \textbf{Client Portraits}. Profiles of clients participating PM+ interventions encompass various aspects including age, gender, occupation, hobbies, health conditions, sources of distress, current mood, and psychiatric symptoms. The number of the portraits is equal to the transcripts.
\item \textbf{Dialogue Generation}. The dialogues are the result of turn-by-turn interactions between the evaluated systems and the 148 clients simulated by the GLM-4 model. We limit the number of turns to 20 as most models typically complete the conversation within this number of turns. 
\item \textbf{Dialogue Assessments}. In this part, we utilize GPT-4 to automatically evaluate the overall quality of the AI psychological counselor's utterances generated during the dialogues. Specifically, by prompting GPT-4 to provide ratings, we assess four key aspects of the AI's responses: logical coherence, professionalism, empathy, and authenticity. 
\end{itemize}

Due to page limitations, we provide the prompts for portrait extraction, the scripts for dialogue generation, and the detailed evaluation metrics in our open-source project.

%% file: 4.experiments.tex
\section{Experiments}

\subsection{Experimental Setup}
We compare SuDoSys with the fine-tuning model CPsyCounX\cite{CPsyCoun}, which has backbone model with 7B parameters. To ensure a fair comparison, we utilize Qwen2-7B-Instruct\cite{qwen} as the base LLM for SuDoSys. Additionally, we employ the stage-unaware prompt stated in \ref{stage_unaware} to instruct Qwen2-7B-Instruct, serving as a baseline method for comparative analysis. For all the models mentioned above, we use the default settings.

In objective evaluation, we employ GPT-4 to assess the interactions between the GLM-4-simulated clients and the evaluated systems. Notably, there are 148 PM+ intervention transcripts and corresponding client portraits, leading to the simulation of 148 clients. Consequently, GPT-4 rates a total of $148\times3=444$ dialogues conducted with these simulated clients. In each evaluation round, GPT-4 reviews the dialogues between the same client and each of the three evaluated systems.

For subjective evaluation, we recruited 20 college students to act as clients and engage in conversations with the evaluated systems. Each participant was asked to select a client portrait extracted from the total of 3,134 multi-turn consultation dialogues in CPsyCounD\cite{CPsyCoun}, and then to interact with each of the three evaluated systems. Following these interactions, each participant rated the three systems according to the same dimensions used by GPT-4. The web interface designed for this subjective evaluation is available on our open-source project.

\subsection{Results and Discussion}

\begin{table}[htbp]
  \centering
  \caption{Objective evaluation results: average ratings given by GPT-4 based on generated dialogue history. Ratings are on an integer scale from 1 to 5, with higher values indicating better performance.}
    \begin{tabular}{c|c|c|c|c}
    \hline
    \textbf{  System  } & \textbf{  Coherence  } & \textbf{  Professionalism  } & \textbf{  Empathy  } & \textbf{  Authenticity  } \\
    \hline
    \textbf{  CPsyCounX  } & \underline{3.9}   & \textbf{4.5} & \textbf{4.4} & \textbf{3.8} \\
    \hline
    \textbf{  Qwen2-7B  } & 3.8   & 3.7   & \underline{4.2}   & \textbf{3.8} \\
    \hline
    \textbf{  SoDuSys  } & \textbf{4} & \underline{4.2}   & 4.1   & \underline{3.7} \\
    \hline
    \end{tabular}%
  \label{tab_automatic}%
\end{table}%

\subsubsection{Objective Evaluation}
As illustrated in Table~\ref{tab_automatic}, SuDoSys outperforms other systems in terms of coherence, demonstrating the efficacy of its stage-aware scheme. Notably, CPsyCounX achieves the highest overall rating, particularly excelling in professionalism and empathy, likely due to its fine-tuning on a substantial corpus of counseling data.

\begin{table}[htbp]
  \centering
  \caption{Subjective evaluation results: ratings given by the 20 students after their interaction with the systems. Ratings follow the same scale as defined in Table~\ref{tab_automatic}.}
    \begin{tabular}{c|c|c|c|c}
    \hline
    \textbf{  System  } & \textbf{  Coherence  } & \textbf{  Professionalism  } & \textbf{  Empathy  } & \textbf{  Authenticity  } \\
    \hline
    \textbf{  CPsyCounX  } & \underline{3.6}   & \textbf{3.7} & \underline{3.5} & \underline{3.7} \\
    \hline
    \textbf{  Qwen2-7B  } & 3.4   & \underline{3.2}   & \textbf{3.8}   & 3.5 \\
    \hline
    \textbf{  SoDuSys  } & \textbf{3.8} & \textbf{3.7}   & \underline{3.5}   & \textbf{3.8} \\
    \hline
    \end{tabular}%
  \label{tab_human}%
\end{table}%

\subsubsection{Subjective Evaluation}
In the subjective evaluation Table~\ref{tab_human}, SuDoSys surpasses other models in coherence with even more pronounced differences than those observed in the automatic evaluation. Furthermore, SuDoSys outperforms CPsyCounX in authenticity. This superiority may be attributed to SuDoSys's ability to manage long-term memories through the incorporation of relevant topics, thereby enhancing its human-like interaction quality.

Overall, leveraging the PM+'s problem managing stages, SuDoSys exhibits a slight advantage in coherence and remains competitive in other dimensions when compared with existing fine-tuning-based counseling models. Moreover, since SuDoSys does not require data for fine-tuning the LLM, it demonstrates a cost-effective and scalable alternative that can potentially reduce the resource burden associated with developing and deploying counseling models while maintaining competitive performance.

%% file: 5.conclusion.tex
\section{Conclusion}
In conclusion, the Structured Dialogue System (SuDoSys) demonstrates a significant improvement in the coherence of counseling dialogues when compared with existing fine-tuning-based models. By leveraging the structured stages outlined in the Problem Management Plus (PM+) guidelines, SuDoSys ensures that the conversations remain coherent and are guided effectively through the counseling process. The system shows competitive performance in professionalism, empathy, and authenticity, as evidenced by both automatic evaluations using GPT-4 and human assessments by college students. These findings highlight the potential of SuDoSys to serve as a valuable tool in bridging the gap in accessible mental health services, particularly in regions lacking specialized professionals. Future work could explore further enhancements to the system, including leveraging real-world counseling datasets to refine the system's performance under complex human-computer interactions.

%% file: bib.tex
%
%
%
\bibliographystyle{splncs04}
%

\bibliography{refs}